\documentclass[conference]{IEEEtran}
\IEEEoverridecommandlockouts

\usepackage{cite}
\usepackage{amsmath,amssymb,amsfonts}
\usepackage{algorithm}
\usepackage{graphicx}
\usepackage{textcomp}
\usepackage{xcolor}
\usepackage{multirow}
\usepackage{colortbl}  
\usepackage{array}
\usepackage{stfloats}
\usepackage{url}
\usepackage{verbatim}
\usepackage{cite}
\usepackage{color}
\usepackage{booktabs}
\usepackage{utfsym}
\usepackage{algpseudocode}

\def\BibTeX{{\rm B\kern-.05em{\sc i\kern-.025em b}\kern-.08em
    T\kern-.1667em\lower.7ex\hbox{E}\kern-.125emX}}
\begin{document}

\title{Distribution-Aware Hadamard Quantization for Hardware-Efficient Implicit Neural Representations}

\author{Wenyong Zhou\textsuperscript{\dag}, Jiachen Ren\textsuperscript{\dag}, Taiqiang Wu, Yuxin Cheng, Zhengwu Liu*, Ngai Wong* \\

Department of Electrical and Electronic Engineering, The University of Hong Kong, Hong Kong SAR \\

$^\dagger$Equal contribution. *Corresponding author:\{zwliu, nwong\}@eee.hku.hk.}

\maketitle

\begin{abstract}
Implicit Neural Representations (INRs) encode discrete signals using Multi-Layer Perceptrons (MLPs) with complex activation functions. While INRs achieve superior performance, they depend on full-precision number representation for accurate computation, resulting in significant hardware overhead. Previous INR quantization approaches have primarily focused on weight quantization, offering only limited hardware savings due to the lack of activation quantization.
To fully exploit the hardware benefits of quantization, we propose DHQ, a novel distribution-aware Hadamard quantization scheme that targets both weights and activations in INRs. Our analysis shows that the weights in the first and last layers have distributions distinct from those in the intermediate layers, while the activations in the last layer differ significantly from those in the preceding layers. Instead of customizing quantizers individually, we utilize the Hadamard transformation to standardize these diverse distributions into a unified bell-shaped form, supported by both empirical evidence and theoretical analysis, before applying a standard quantizer. To demonstrate the practical advantages of our approach, we present an FPGA implementation of DHQ that highlights its hardware efficiency. Experiments on diverse image reconstruction tasks show that DHQ outperforms previous quantization methods, reducing latency by 32.7\%, energy consumption by 40.1\%, and resource utilization by up to 98.3\% compared to full-precision counterparts.
\end{abstract}
\section{Introduction}
\label{sec:intro}
Implicit Neural Representations (INRs) have emerged as a powerful approach for encoding signals~\cite{liu2024finer, xie2023diner, zhang2024ntinr}. Unlike traditional discrete representations, INRs leverage neural networks, typically Multi-Layer Perceptrons (MLPs), to learn continuous mappings from coordinates to signal values, enabling high-fidelity reconstruction of images, videos, and 3D scenes~\cite{liasmr, saragadam2023wire}. To capture the finer details in input signals, INRs leverage more complex activation functions, such as sinusoidal and wavelet functions, instead of the ReLU function that has been widely adopted in other Deep Neural Networks (DNNs)~\cite{sitzmann2020siren, saragadam2023wire}. Accurate computations in INRs is critical for achieving the desired model performance, thus it generally relies on 32-bit full-precision representations of weights and activations. However, this high precision leads to substantial computational and memory requirements, limiting INR deployment on resource-constrained hardware platforms~\cite{TernaryNeRF}.

Quantization reduces computational and memory access overhead during model execution by replacing high-precision floating-point values with low-precision representations for weights, activations, and gradients. Compared to 32-bit floating-point formats, 8-bit integer formats reduce storage requirements by 75\%, lower energy consumption by an order of magnitude, and decrease circuit area by two orders of magnitude~\cite{low_bit_train}. 

\begin{figure}[!t]
\centering
\includegraphics[scale=0.25]{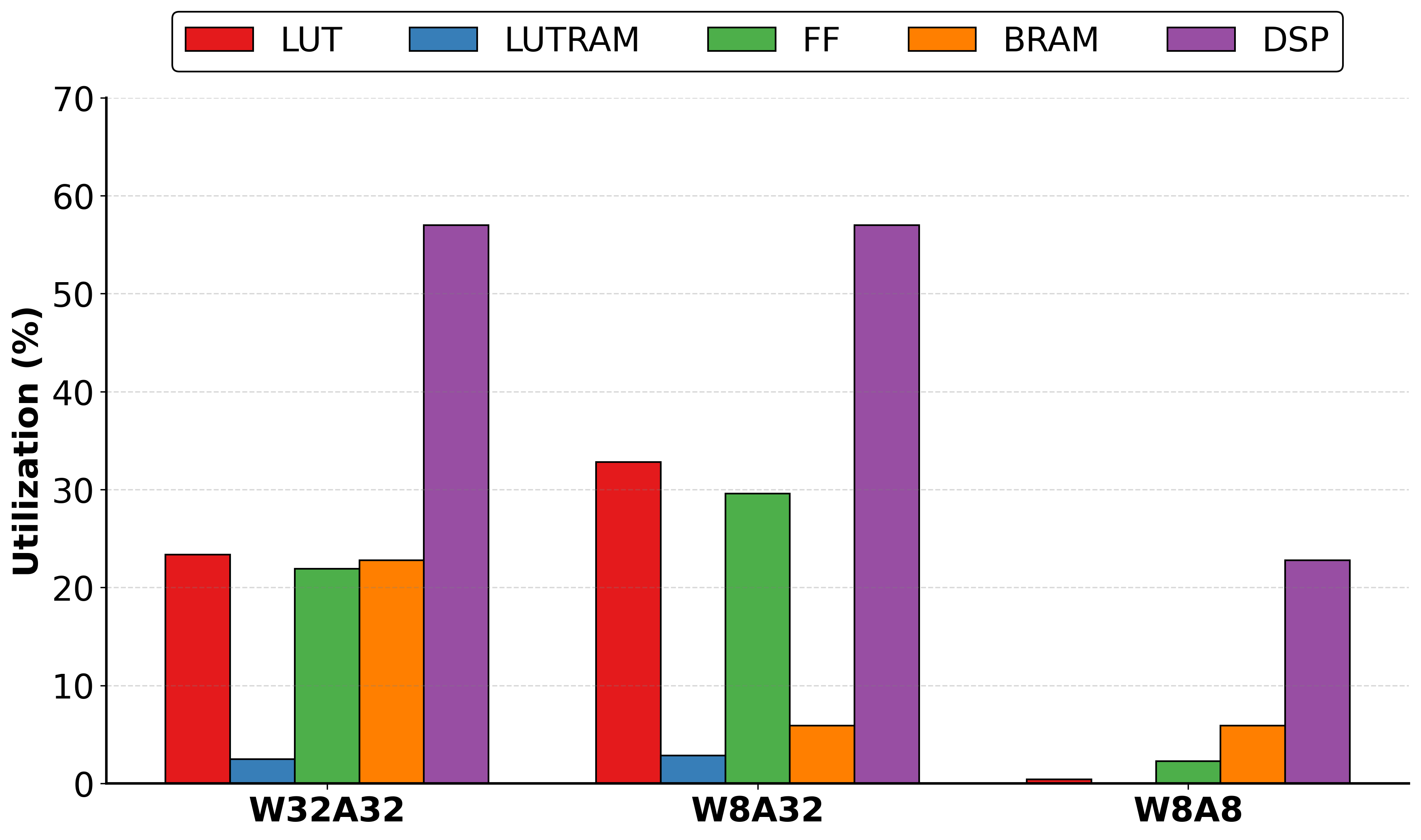}
\caption{Resource utilization comparison across different quantization configurations for SIREN model implementation. The plot shows hardware resource utilization (\%) of LUT, LUTRAM, FF, BRAM, and DSP for differnet bit width configurations (W32A32, W8A32, W8A8). } 
\label{fig:hw_cost}
\vspace{-0.2cm}
\end{figure}

Previous works highlight the potential of INR quantization. For instance,~\cite{gordon2023quantizing} employed K-means clustering to assign bit-widths across all layers based on a detailed analysis of quantization ranges. While their approach effectively minimizes performance degradation under low bit-width, it has several drawbacks. First, the K-means algorithm requires a computationally intensive iterative process for each quantization step and depends on manually set hyperparameters to determine the stopping threshold, which can result in suboptimal compression outcomes. Second, the practical hardware savings of their method are limited due to the absence of activation quantization. Expensive high-precision computations are still required to multiply low-precision weights with high-precision activations. We present the hardware costs of floating-point (W32A32), mixed-precision (W8A32), and fixed-point (W8A8) calculations on an Field Programmable Gate Arrays (FPGA) platform in Figure~\ref{fig:hw_cost}. It is evident that while mixed-precision calculations significantly reduce storage requirements (particularly for BRAM), they consume more resources than full-precision computations due to the overhead of data conversion. Specifically, LUT utilization increases to 32.79\%, while DSP usage remains high at 56.98\%. In contrast, fixed-point calculations offer substantial hardware benefits, reducing LUT utilization by 98.3\%, LUTRAM by 99.6\%, and DSP usage by 60.0\%.
\begin{figure}[!t]
\centering
\includegraphics[scale=0.4]{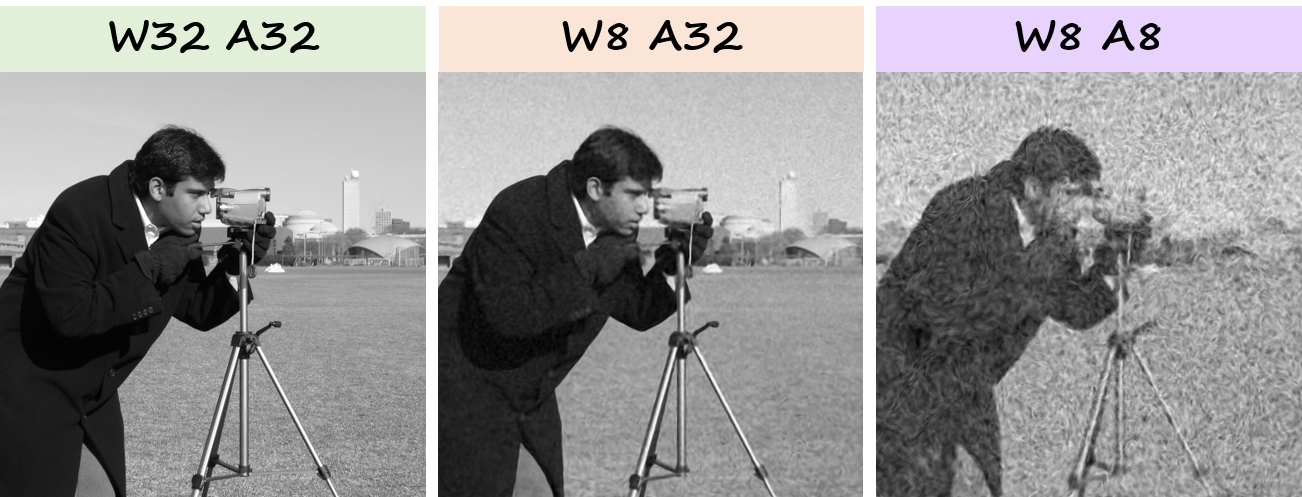}
\caption{Visualization of weight and activation quantization effects on grayscale image reconstruction using the SIREN model~\cite{sitzmann2020siren}. Results are shown for full precision (W32A32), weight-only quantization (W8A32), and weight and activation quantization (W8A8) using a standard uniform 8-bit quantizer.} 
\label{fig:motivation}
\vspace{-0.2cm}
\end{figure}

However, activation quantization poses a significant challenge due to its high dynamic range and complex distribution, which is particularly critical for INRs given the sensitive nature of the reconstruction task~\cite{lbt}. As shown in Figure~\ref{fig:motivation}, while weight-only quantization (W8A32) maintains reasonable image quality, joint weight and activation quantization (W8A8) leads to substantial quality degradation in the reconstructed image.

To address these challenges, we propose DHQ, a novel distribution-aware Hadamard quantization method for joint weight and activation quantization in INRs. DHQ utilizes the Hadamard transformation to reshape diverse distributions into bell-shaped forms, enabling the use of a simple quantizer for efficient quantization. In summary, our key contributions are threefold:
\begin{itemize}
    \item We observe that weights in the first and last layers exhibit uniform or bimodal distributions, which are distinct from the bell-shaped distributions found in the intermediate layers. Similarly, activations in the last layer show a more centrally concentrated distribution, while those in preceding layers display U-shaped distributions with explicit bounds.
    \item We propose DHQ, a novel distribution-aware Hadamard quantization method for both weights and activations in INRs. DHQ leverages the Hadamard transformation to standardize diverse weight and activation distributions into a unified bell-shaped form, enabling the use of a standard quantizer. Moreover, we present an FPGA implementation of DHQ to demonstrate its practical hardware savings.
    \item Extensive experiments on image reconstruction tasks, including grayscale, color, and biomedical images, demonstrate that DHQ achieves reconstruction quality comparable to full-precision models, while reducing latency by 32.7\%, improving energy efficiency by 40.1\%, and lowering resource utilization by up to 98.3\% compared to full-precision counterparts.
\end{itemize}
\section{Related Work}
\label{sec:related}
\subsection{INR}
INR is a neural network-based framework that learns a continuous function $f_\theta(\textbf{x}) \rightarrow \textbf{y}$, mapping coordinate inputs \textbf{x} to feature outputs \textbf{y} through learnable parameters $\theta$~\cite{mildenhall2020nerf,dupont2021coi}. This powerful representation has found applications across diverse domains, from 3D scene reconstruction with NeRF and its variants~\cite{mescheder2019occupancynetworks,park2019deepsdf}, to multimedia processing including audio~\cite{sitzmann2020siren,lu2021compressive} and video~\cite{martel2021acorn,xie2023diner}, and various spatial representations such as textures~\cite{texture}, geometry~\cite{geometric}, and topological structures~\cite{vn}.

\subsection{Quantization}
Quantization has emerged as a crucial technique for reducing the memory and computational costs of neural networks by converting floating-point weights and activations to lower-precision representations. Traditional quantization methods can be broadly categorized into uniform and non-uniform quantization~\cite{gordon2023quantizing}. Uniform quantization maps a floating-point value $x$ to its quantized version $x_q$ using a linear transformation:
\begin{equation}
    x_q = \text{round}(\frac{x - x_{\text{min}}}{x_{\text{max}} - x_{\text{min}}} \cdot (2^b - 1)) \cdot s + z,
\end{equation}
where $b$ is the bit width, $s$ is the scaling factor, and $z$ is the zero point. While uniform quantization offers hardware-friendly implementation, it may sacrifice accuracy when dealing with skewed distributions. 

Non-uniform quantization adapts the quantization intervals to the underlying data distribution. A common approach is logarithmic quantization, which maps values using a logarithmic scale. This approach achieves better accuracy for values of different magnitudes while maintaining relatively simple hardware implementation, as it can be realized through bit-shift operations.

Recent works have highlighted the challenges in neural network activation quantization~\cite{SRUQ,FAST}. While weight quantization primarily deals with static parameters, activation quantization must handle dynamic ranges during inference. Traditional deterministic rounding methods often lead to significant accuracy degradation due to quantization error accumulation. To address this, stochastic rounding has emerged as an effective approach~\cite{SRUQ,FAST}, where a value $x$ is probabilistically rounded to either its floor or ceiling:
\begin{equation}
    Q(x) = \begin{cases}
    \lfloor x \rfloor & \text{with probability } 1 - (x - \lfloor x \rfloor) \\
    \lceil x \rceil & \text{with probability } (x - \lfloor x \rfloor)
    \end{cases}
\end{equation}
This unbiased estimator helps preserve the expected value during training and reduces gradient approximation errors. Activation quantization in INRs faces challenges beyond inference dynamic range. The reconstruction tasks inherently contain less redundancy than classification tasks, making them more vulnerable to errors. Furthermore, INRs' complex activation functions heighten their sensitivity to even small quantization errors.
\section{Methodology}
\label{sec:method}
In this section, we examine the layer-wise distribution characteristics of weights and activations in a SIREN model. Building on our observations, we present a distribution-aware quantization scheme for both weights and activations, along with its practical realization through an FPGA-based hardware architecture.
\subsection{Weight and Activation Distribution of INR}
\begin{figure}[!t]
\centering
\includegraphics[scale=0.33]{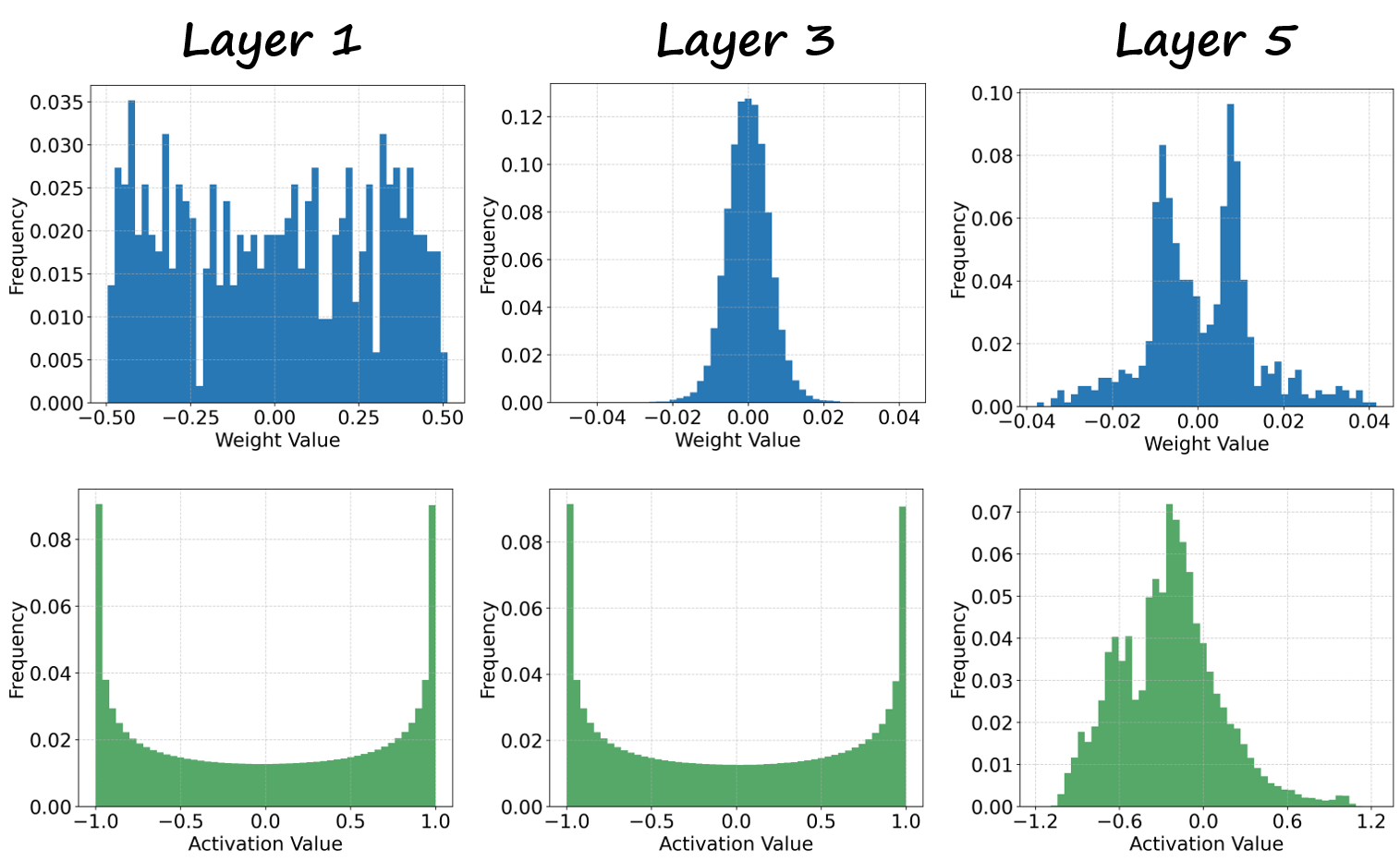}
\caption{Visualization of weight and activation distributions across different layers (1, 3, and 5) of a SIREN model. Top row (blue) shows weight distributions, while bottom row (green) shows activation distributions for each layer during \texttt{Kodak02} image reconstruction.} 
\label{fig:distribution}
\vspace{-0.2cm}
\end{figure}

The distribution of high-precision values plays a crucial role in quantization, as it directly impacts the effectiveness of converting these values to lower-precision formats. For uniformly distributed values, uniform quantization, which divides the range into equal intervals, can be highly effective. However, real-world data often follows non-uniform distributions, such as normal or skewed patterns. In these scenarios, uniform quantization becomes inefficient, as it allocates resources equally across the entire range, regardless of the frequency of occurrences. Instead, non-uniform quantization, which dedicates finer resolution to regions with higher data density, often yields better performance at the same bit width. Figure~\ref{fig:distribution} illustrates the distributions of weights and activations across three representative layers (1, 3, and 5) of a SIREN neural network trained for color image reconstruction.

The weight distributions reveal distinct characteristics across layers. The first layer displays a relatively uniform distribution from $-0.5$ to $0.5$, while the third layer shows a more concentrated, bell-shaped distribution centered around zero, with a narrower range ($-0.04$ to $0.04$). Although the second and fourth layer distributions are omitted, they resemble the third layer due to similar input and output dimensions. The fifth layer's weight distribution, in contrast, exhibits a bell-shaped pattern with two distinct peaks.

The activation distributions also display intriguing patterns. In the first and third layers, activations follow a U-shaped pattern ranging from $-1$ to $1$, resulting from the sine activation function in the SIREN architecture, which tends to push values towards the extremes of the range due to its periodic nature. However, the activation distribution of the fifth layer is different, showing a bell-shaped distribution centered around zero. This is because the fifth layer is a linear output layer without the sine activation function, designed to directly produce the final pixel values for image reconstruction, resulting in a more conventional distribution pattern. Thus, special attention is needed when quantizing activations in the last layer.

These observed distribution patterns suggest specific quantization strategies for SIREN networks. The broader, more uniform weight distribution in the first layer indicates that uniform quantization could be effective there, whereas the concentrated distributions in intermediate layers point towards non-uniform quantization as an optimal approach. For activations, the U-shaped distributions in earlier layers highlight the non-uniform utilization of the range, with concentrations at the extremes. In contrast, the bell-shaped activation distribution in the fifth layer resembles more traditional patterns, necessitating a different quantization approach for this final layer.
\subsection{Distribution-aware Hadamard Quantization}
As demonstrated in the previous section, the weights and activations in INRs exhibit distinct distribution patterns across different layers. The weight distributions vary notably: the first layer exhibits a relatively uniform distribution, the intermediate layers show concentrated bell-shaped distributions, and the final layer reveals a distinctive bimodal pattern. Similarly, the activation distributions are highly characteristic, with hidden layers displaying U-shaped patterns due to the sine activation function, while the output layer follows a conventional bell-shaped distribution. These layer-specific properties suggest that a distribution-aware quantization strategy is essential for achieving more effective compression.

The uniformly distributed weights in the first layer and the U-shaped, symmetric, zero-mean distributed activations are well-suited to classical uniform quantization with equally spaced levels. In contrast, unevenly distributed weights and activations, particularly the bimodal weight distribution in the final layer, demand specialized considerations. Tailoring the quantization process to the specific distribution patterns of weights and activations can significantly enhance the fidelity of compressed models. However, this approach introduces additional complexity, as it necessitates more sophisticated analysis and the design of custom quantizers for each layer.

Instead of designing different quantizers for each layer, we propose applying Hadamard transformations to weights and activations, allowing the use of a single quantizer across all layers~\cite{hadamard}. The key insight lies in utilizing Hadamard transformations tailored to the unique distributions of each layer, standardizing them into a unified distribution before applying a common quantization strategy. This approach simplifies the quantization process while preserving its effectiveness. Our idea is inspired by the design of SIREN, where the sine activation function transforms various input distributions into similar U-shaped output distributions. 
\begin{figure}[!t]
\centering
\includegraphics[scale=0.4]{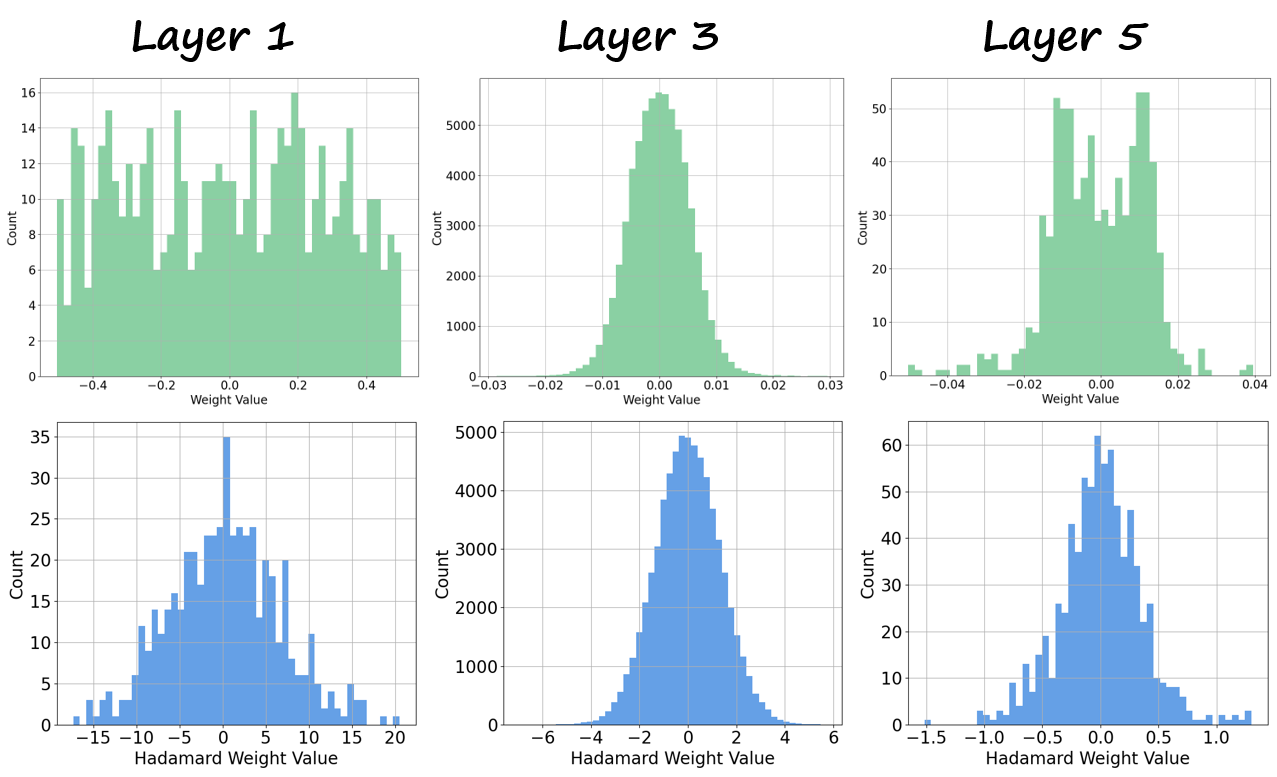}
\caption{Weight distributions before (top row) and after (bottom row) applying the Hadamard transformation for different layers of the INR model. After the Hadamard transformation, all layers are unified into a bell-shaped distribution.}
\label{fig:hadamard}
\vspace{-0.2cm}
\end{figure}

The Hadamard transformation is a linear operation that uses an orthogonal matrix known as the Hadamard matrix to transform data. The Hadamard matrix $H_n$, of size $n \times n$, is defined recursively for dimensions that are powers of two:
\begin{equation}
H_1 = \begin{bmatrix} 1 \end{bmatrix} \quad
H_{2n} = \begin{bmatrix} H_n & H_n \\ H_n & -H_n \end{bmatrix}
\end{equation}
The rows of $H_n$ are orthogonal, satisfying the property $H_n^T H_n = nI$, where $I$ is the identity matrix. When applied to a vector $X \in \mathbb{R}^n$, the Hadamard transform redistributes its values while preserving its total energy. For 2D matrices $W \in \mathbb{R}^{m \times n}$, the transformation is extended as:
\begin{equation}
W' = H_m \cdot W \cdot H_n,
\end{equation}
where $H_m$ and $H_n$ are Hadamard matrices corresponding to the number of rows and columns, respectively. When the Hadamard transformation is applied to a 2D matrix $W$, it has the intriguing effect of shaping the distribution of its entries into a bell-shaped, Gaussian-like distribution. Mathematically, each element $W'_{ij}$ in the transformed matrix is expressed as:
\begin{equation}
W'_{ij} = \frac{1}{\sqrt{mn}} \sum_{k=1}^m \sum_{l=1}^n H_{ik} W_{kl} H_{lj}
\label{eqn:element}
\end{equation}
where $H_{ik}$ and $H_{lj}$ are entries from the Hadamard matrices for rows and columns, respectively. Equation~\ref{eqn:element} shows that $W'_{ij}$ is a linear combination of all elements in $W$, weighted by the orthogonal components of the Hadamard matrices.

The transformation preserves the total energy of $W$ (measured by its Frobenius norm, $\|W\|_F$) due to the orthogonality of the Hadamard matrix, ensuring $\|W'\|_F = \|W\|_F$. However, the structured mixing of entries redistributes the energy across the matrix, altering the statistical distribution of its values. Importantly, the entries of $W'$ become sums of many independent contributions $H_{ik} W_{kl} H_{lj}$. The statistical behavior of the transformed matrix $W'$ can be understood through the Central Limit Theorem. If the entries of $W$ are drawn from a distribution with finite variance, the sums $W'_{ij}$ converge to a Gaussian distribution as the dimensions $m$ and $n$ grow large. Specifically, let the mean of the entries of $W$ be $\mu$. Since the Hadamard transform is a linear operation, the mean of $W'$ remains unchanged:
\begin{equation}
\mathbb{E}[W'_{ij}] = \mathbb{E}\left[\frac{1}{\sqrt{mn}} \sum_{k=1}^m \sum_{l=1}^n H_{ik} W_{kl} H_{lj}\right] = \mu
\end{equation}
Suppose the variance of the entries of $W$ is $\sigma^2$, the variance of $W'_{ij}$ becomes:
\begin{equation}
\text{Var}(W'_{ij}) = \frac{\sigma^2}{mn}
\end{equation}
This result indicates that as the size of the matrix increases, the entries of $W'$ become more tightly clustered around the mean. Consequently, for large $m$ and $n$, the entries of $W'$ approximate a Gaussian distribution:
\begin{equation}
W'_{ij} \sim \mathcal{N}(\mu, \sigma^2 / mn).
\end{equation}
As illustrated in Figure~\ref{fig:hadamard}, the initially distinct weight distributions of the first and last layers are transformed into a unified bell-shaped distribution after the Hadamard transform. For the weights in the intermediate layers, the transformation produces a distribution that closely resembles a Gaussian shape. This ability to standardize diverse distributions reduces variability and introduces regularity into the data, enabling the application of a simple quantizer for efficiently handling data with different original distributions.
\subsection{FPGA Design and Implementation}
\begin{figure}[!t]
\centering
\includegraphics[scale=0.3]{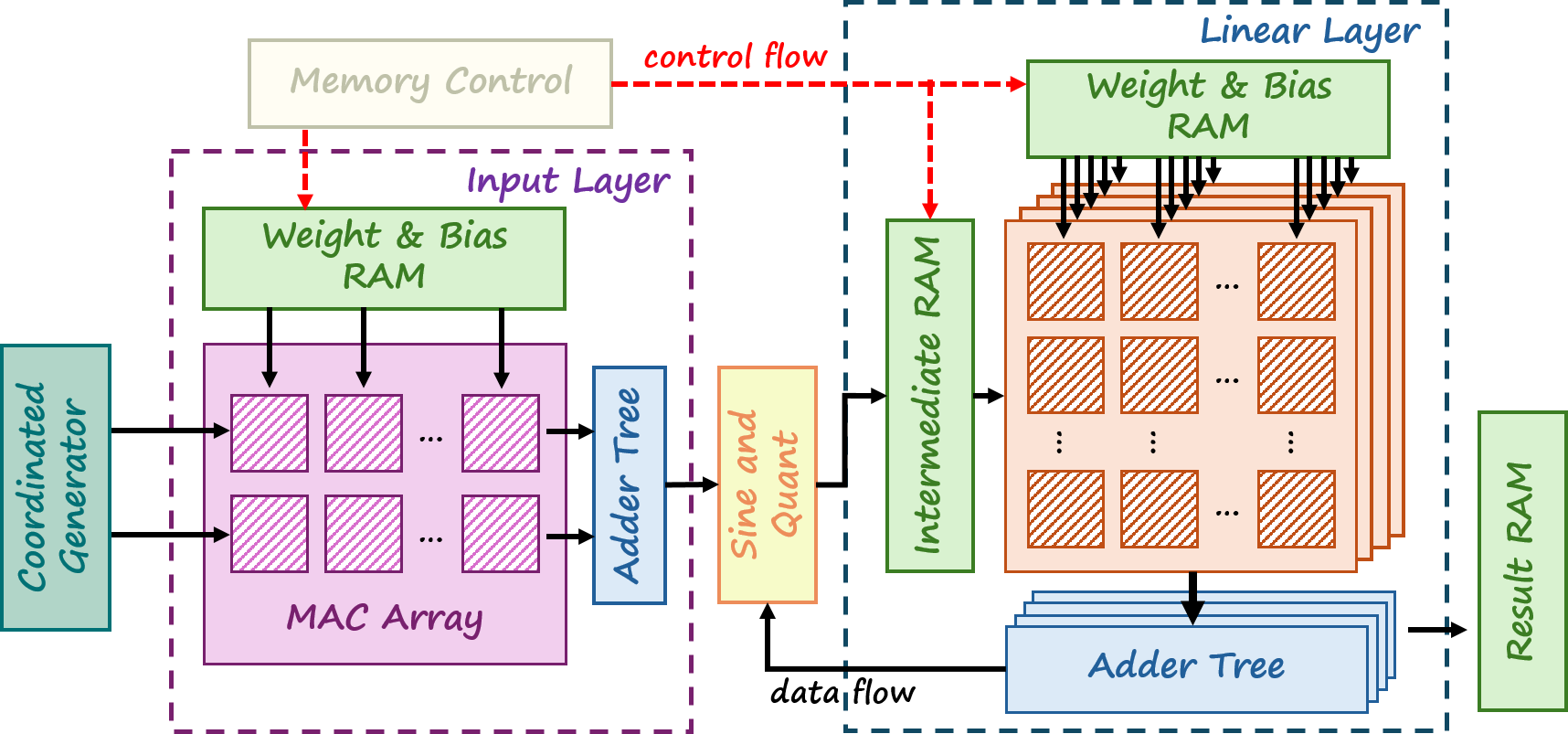}
\caption{Hardware architecture of the proposed INR accelerator on FPGA, consisting of specialized input and linear layers with 256 parallel MAC arrays, dedicated RAM blocks for weights and intermediate results, and pipelined processing units for coordinate-to-pixel generation.}
\label{fig:fpga}
\vspace{-0.2cm}
\end{figure}
Figure~\ref{fig:fpga} presents the architecture of a specialized FPGA-based accelerator designed for INR tasks, integrating pipelined execution and parallel computation to achieve efficient coordinate-to-image generation. The system is structured around three core modules—input layer, linear layer, and output layer—coordinated by a centralized memory control unit. The input layer processes spatial coordinates through a compact MAC array and a coordinate-activated generator, mapping them into high-dimensional feature vectors. A sine activation function injects nonlinearity into these features, critical for capturing high-frequency spatial patterns. To optimize interlayer pipelining, the input layer employs a reduced-scale MAC array and dedicated Weight and Bias RAM blocks, ensuring rapid parameter access while minimizing logic overhead. This design allows concurrent prefetching of data for subsequent layers, maximizing throughput.

The linear layer serves as the computational backbone, leveraging a large-scale parallel MAC array and an adder tree to perform high-density arithmetic operations. Each MAC unit processes weighted activations using parameters stored in Weight and Bias RAM, while the adder tree aggregates partial sums into intermediate results buffered in Intermediate RAM. Between layers, a dynamic quantization module adaptively reduces activation precision, enabling seamless transitions between quantization strategies such as W8A32 and W8A8. The linear layer’s parallelism is balanced with fine-grained pipelining to avoid memory bandwidth bottlenecks, ensuring sustained utilization of computational resources.

In contrast to the linear layer, the output layer omits activation functions to streamline final result generation. The accelerator’s design emphasizes flexibility and scalability, with parameterized components such as MAC array size and RAM depth tailored to accommodate varying network complexities. These optimizations, combined with dynamic quantization support, position the system as a versatile solution for edge-based INR applications, where adaptability and hardware efficiency are paramount. The integration of parallel computation, pipelined execution, and configurable precision modules exemplifies a holistic approach to deploying neural representations on resource-constrained FPGA platforms.

\section{Experiments}
\label{sec:experiment}
In this section, we evaluate DHQ against previous INR quantization approaches across various reconstruction tasks. In addition, we assess the hardware efficiency of our method through FPGA simulations to demonstrate the practical benefits of DHQ.
\subsection{Experiment Setup}
We evaluate our method using two INR architectures: SIREN as the baseline model and WIRE, a state-of-the-art INR model featuring wavelet activation functions, to demonstrate the generalization capability of DHQ. All quantization experiments are performed on a NVIDIA GeForce RTX 3090 GPU with 24GB memory. For hardware evaluation, we conduct experiments on a Xilinx Virtex UltraScale+ HBM VCU128 FPGA platform.

\subsection{Experiment Results}
We compared our method (DHQ) with several previous approaches on the {\tt cameraman} image reconstruction task using both SIREN~\cite{sitzmann2020siren} and WIRE~\cite{saragadam2023wire} architectures. Uniform quantization divides the value range from the minimum and maximum values into equal intervals. In contrast, the K-Means method proposed in~\cite{gordon2023quantizing} is a heuristic non-uniform quantization. As shown in Table~\ref{tab:quant_comparison}, our DHQ method performs competitively compared to the baselines in terms of both peak signal-to-noise ratio (PSNR) and structural similarity index (SSIM), while utilizing a lower bit width. For the SIREN architecture, our method achieves a PSNR of 31.57 dB and an SSIM of 0.87 with only 8/8 bits, performing better than uniform quantization (29.65 dB PSNR, 0.84 SSIM) with the same bit width, though slightly lower than K-Means which achieves 31.98 dB PSNR and 0.88 SSIM with 8/8 bits. Similarly, for the WIRE architecture, DHQ achieves a PSNR of 31.94 dB and an SSIM of 0.88, outperforming uniform quantization (30.54 dB PSNR, 0.86 SSIM) while using the same 8/8 bit width.

\begin{figure}[!t]
\centering
\includegraphics[scale=0.38]{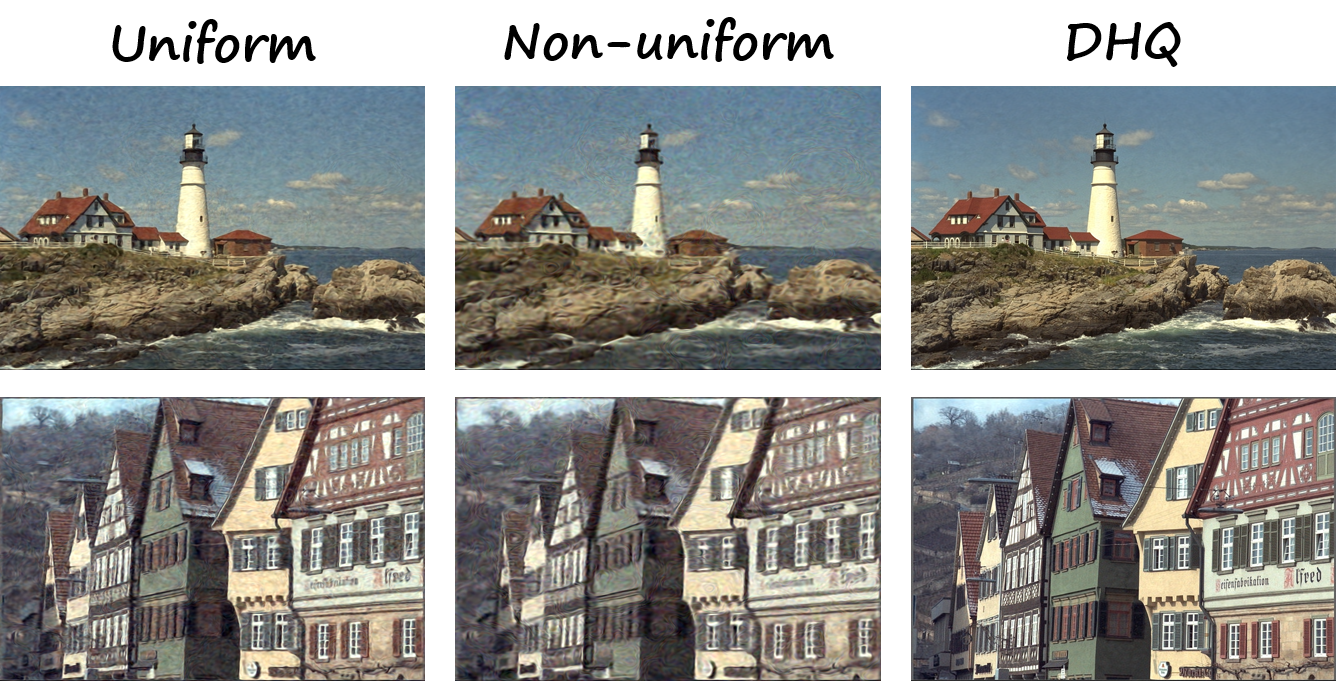}
\caption{Comparison between classical uniform quantization, non-uniform quantization, and our proposed DHQ method for 8-bit weight and activation quantization on \textbf{natural images}.} 
\label{fig:kodak}
\vspace{-0.2cm}
\end{figure}

\begin{figure}[!t]
\centering
\includegraphics[scale=0.38]{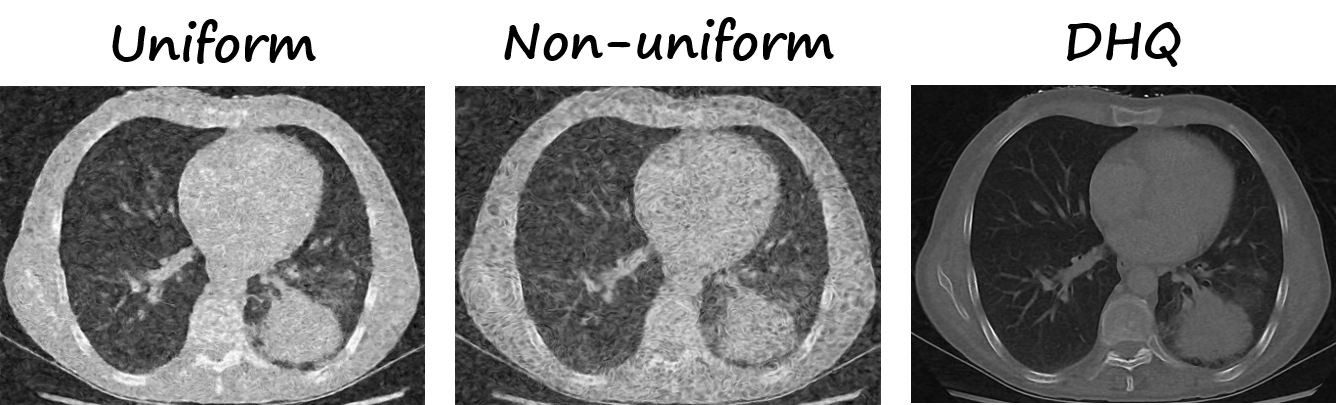}
\caption{Comparison between classical uniform quantization, non-uniform quantization, and our proposed DHQ method for 8-bit weight and activation quantization on \textbf{CT images}.} 
\label{fig:ct}
\vspace{-0.2cm}
\end{figure}

We present visual comparisons between different quantization methods on both natural images (Kodak dataset) and medical images (CT scans) in Figures~\ref{fig:kodak} and~\ref{fig:ct}. For the Kodak images, although classical uniform and non-uniform quantization reconstruct the main components in the images, they fail to capture the detailed structures of the lighthouse, surrounding buildings, and architectural features. However, our DHQ method successfully preserves fine details while maintaining natural contrast, particularly visible in the texture of the lighthouse walls, the intricate architectural elements of the buildings, and the subtle variations in the sky and landscape. In the case of CT images, the differences between methods become more apparent. The classical uniform and non-uniform quantization methods depict the contours of key anatomical structures but contain obvious noise in the image resulting in low contrast, while our DHQ provides better contrast with clearer boundaries between different tissue structures.
\begin{table}[!t]
\renewcommand{\arraystretch}{1.3}
\caption{Comparison of Different INR Quantization Methods for Reconstruction of the Cameraman Image.}
\definecolor{highlight}{HTML}{e2f0cb}
\centering
\begin{tabular}{l c c c}
\toprule
\textbf{Model / Method} & \textbf{W/A Bit Width} & \textbf{PSNR (dB)} & \textbf{SSIM} \\
\midrule
\multicolumn{4}{l}{\textbf{SIREN}~\cite{sitzmann2020siren}} \\
\quad Full-precision                               & 32 / 32            & 32.13         & 0.89 \\
\quad K-Means~\cite{gordon2023quantizing}   & 8 / 8            & 30.98         & 0.84 \\
\quad Uniform                               & 8 / 8            & 29.65         & 0.82 \\
\rowcolor{highlight}\quad DHQ (Ours)       & 8 / 8             & 31.57         & 0.87 \\
\midrule
\multicolumn{4}{l}{\textbf{WIRE}~\cite{saragadam2023wire}} \\
\quad Full-precision                                 & 32 / 32            & 33.15         & 0.89 \\
\quad K-Means~\cite{gordon2023quantizing}   & 8 / 8            & 31.07         & 0.87 \\
\quad Uniform                               & 8 / 8            & 30.54         & 0.84 \\
\rowcolor{highlight}\quad DHQ (Ours)       & 8 / 8              & 31.94             & 0.88 \\
\bottomrule
\end{tabular}
\label{tab:quant_comparison}
\end{table}
 
\begin{table}[!t]
\renewcommand{\arraystretch}{1.3}
\caption{Hardware costs of SIREN models with different weight and activation quantization.}
\centering
\begin{tabular}{lccc}
\toprule
\multirow{2}{*}{\textbf{Hardware Costs}} & \multicolumn{3}{c}{\textbf{Weight and Activation Bit Width}} \\
\cmidrule(lr){2-4}
& \textbf{W32 A32} & \textbf{W8 A32} & \textbf{W8 A8} \\ 
\midrule
Latency (cycle)                     & 1699      & 1734          & 1143 \\
Power (W)                           & 6.758     & 7.979         & 4.051 \\
Resource Utilization                &           &               & \\
\quad Look-Up Table                 & 304252    & 427537        & 5081 \\
\quad Look-Up Table RAM             & 14857     & 17041         & 60 \\
\quad Flip-Flop                     & 570781    & 771026        & 59489 \\
\quad Digital Signal Processor      & 5142      & 5142          & 2056 \\
\quad Block RAM                     & 459.5     & 119           & 119 \\
\bottomrule
\end{tabular}
\label{tab:hw_result}
\end{table}
\subsection{Hardware Evaluation}
To quantify the hardware cost of different quantization schemes, we evaluate the hardware costs of SIREN models across different quantization schemes (W32A32, W8A32, and W8A8) in Table~\ref{tab:hw_result}. Our analysis reveals that weight-only quantization (W8A32) shows notable storage efficiency, reducing Block RAM usage from 459.5 to 119, a 74.1\% reduction. However, this approach leads to increased hardware overhead, with latency rising from 1699 to 1734 cycles (+2.1\%) and power consumption increasing from 6.758W to 7.979W (+18.1\%). The DSP utilization remains constant at 5142 units, indicating no improvement in computational efficiency. In contrast, implementing DHQ (W8A8) yields comprehensive improvements: latency decreases significantly to 1143 cycles (32.7\% reduction), power consumption optimizes to 4.051W (40.1\% decrease), DSP utilization drops to 2056 units (60\% reduction), and LUT requirements decrease from 304252 to 5081 (98.3\% reduction), with LUTRAM usage becoming minimal at 60 units.

The dramatic reduction in resource utilization while maintaining the same BRAM footprint as W8A32 suggests that DHQ achieves an optimal balance between hardware efficiency and model performance. The 60\% reduction in DSP utilization creates opportunities for increased parallelization, potentially enabling further latency improvements. These results demonstrate that DHQ successfully optimizes both computational complexity and memory access patterns, making it particularly suitable for hardware-efficient INR implementations.
\section{Conclusion}
\label{sec:conclusion}

In this paper, we introduced DHQ, a novel distribution-aware Hadamard quantization framework designed for INRs. Based on our observation that weights and activations across different layers exhibit distinct distributions in INRs, we leverage the Hadamard transformation to standardize these diverse distributions into a unified bell-shaped form, enabling the use of a standard quantizer. Additionally, we present an FPGA implementation of DHQ to demonstrate its practical effectiveness. Experiments on image reconstruction tasks show that DHQ achieves reconstruction quality comparable to full-precision models, while reducing latency by 32.7\%, improving energy efficiency by 40.1\%, and reducing resource utilization by up to 98.3\% compared to full-precision counterparts.
\section*{Acknowledgment}
This work was supported in part by the Theme-based Research Scheme (TRS) project T45-701/22-R, National Natural Science Foundation of China (62404187) and the General Research Fund (GRF) Project 17203224, of the Research Grants Council (RGC), Hong Kong SAR.

\bibliographystyle{IEEEbib}
\bibliography{icme2025references}

\end{document}